\documentclass[final, 5p,times,twocolumn, preprint]{elsarticle}

\usepackage{amsmath, amssymb, amsfonts}
\usepackage{algorithm}
\usepackage{algpseudocode}
\usepackage{enumerate}
\usepackage{tabularx}
\usepackage{textcomp}
\usepackage{amsthm}
\usepackage{booktabs}
\usepackage{multirow}
\usepackage{ulem}
\usepackage{tabularx}
\usepackage{threeparttable}
\usepackage{caption}
\usepackage{makecell}
\usepackage[switch]{lineno}
\usepackage{color}
\usepackage[table]{xcolor} 
\usepackage{url}
\usepackage{bbding}
\usepackage{wrapfig}

\newcolumntype{L}[1]{>{\raggedright\arraybackslash}p{#1}}
\newcolumntype{C}[1]{>{\centering\arraybackslash}p{#1}}
\newcolumntype{R}[1]{>{\raggedleft\arraybackslash}p{#1}}
\newcommand{\settablefont}{\fontsize{7.5}{8.6}\selectfont}

\newcommand{\etc}{\textit{etc}}

\journal{Neurocomputing}

\begin{document}
\begin{frontmatter}



\title{WP-CrackNet: A Collaborative Adversarial Learning Framework for End-to-End Weakly-Supervised Road Crack Detection}

\author[label1]{Nachuan Ma}\ead{2111481@tongji.edu.cn}
\author[label1]{Zhengfei Song}\ead{2151094@tongji.edu.cn}
\author[label1]{Qiang Hu}\ead{2252974@tongji.edu.cn}
\author[label2]{Xiaoyu Tang}\ead{tangxy@scnu.edu.cn}
\author[label3]{Chengxi Zhang}\ead{cxzhang@jiangnan.edu.cn}
\author[label1]{Rui Fan\corref{cor}}\ead{rui.fan@ieee.org}
\author[label4]{Lihua Xie}\ead{elhxie@ntu.edu.sg}

\affiliation[label1]{organization= {College of Electronic and Information Engineering, Shanghai Research Institute for Intelligent Autonomous Systems, Tongji University},
            country={China}}    
\affiliation[label2]{organization= {School of Electronics and Information Engineering, and Xingzhi College, South China Normal University},
            country={China}} 
\affiliation[label3]{organization= {School of Internet of Things Engineering, Jiangnan University},
            country={China}}  
\affiliation[label4]{organization= {School of Electrical and Electronic Engineering, Nanyang Technological University},
            country={Singapore}} 
            

\cortext[cor]{R. Fan is the corresponding author.}

\begin{abstract}
Road crack detection is essential for intelligent infrastructure maintenance in smart cities. To reduce reliance on costly pixel-level annotations, we propose WP-CrackNet, an end-to-end weakly-supervised method that trains with only image-level labels for pixel-wise crack detection. WP-CrackNet integrates three components: a classifier generating class activation maps (CAMs), a reconstructor measuring feature inferability, and a detector producing pixel-wise road crack detection results. During training, the classifier and reconstructor alternate in adversarial learning to encourage crack CAMs to cover complete crack regions, while the detector learns from pseudo labels derived from post-processed crack CAMs. This mutual feedback among the three components improves learning stability and detection accuracy. To further boost detection performance, we design a path-aware attention module (PAAM) that fuses high-level semantics from the classifier with low-level structural cues from the reconstructor by modeling spatial and channel-wise dependencies. Additionally, a center-enhanced CAM consistency module (CECCM) is proposed to refine crack CAMs using center Gaussian weighting and consistency constraints, enabling better pseudo-label generation.
We create three image-level datasets and extensive experiments show that WP-CrackNet achieves comparable results to supervised methods and outperforms existing weakly-supervised methods, significantly advancing scalable road inspection. The source code package and datasets are available at https://mias.group/WP-CrackNet/.     
\end{abstract}

\begin{keyword}
Weakly-supervised, road cracks, deep learning, computer vision.
\end{keyword}

\end{frontmatter}


\section{Introduction}

\begin{figure}[!t]
	\begin{center}
		\centering
		\includegraphics[width=0.48\textwidth]{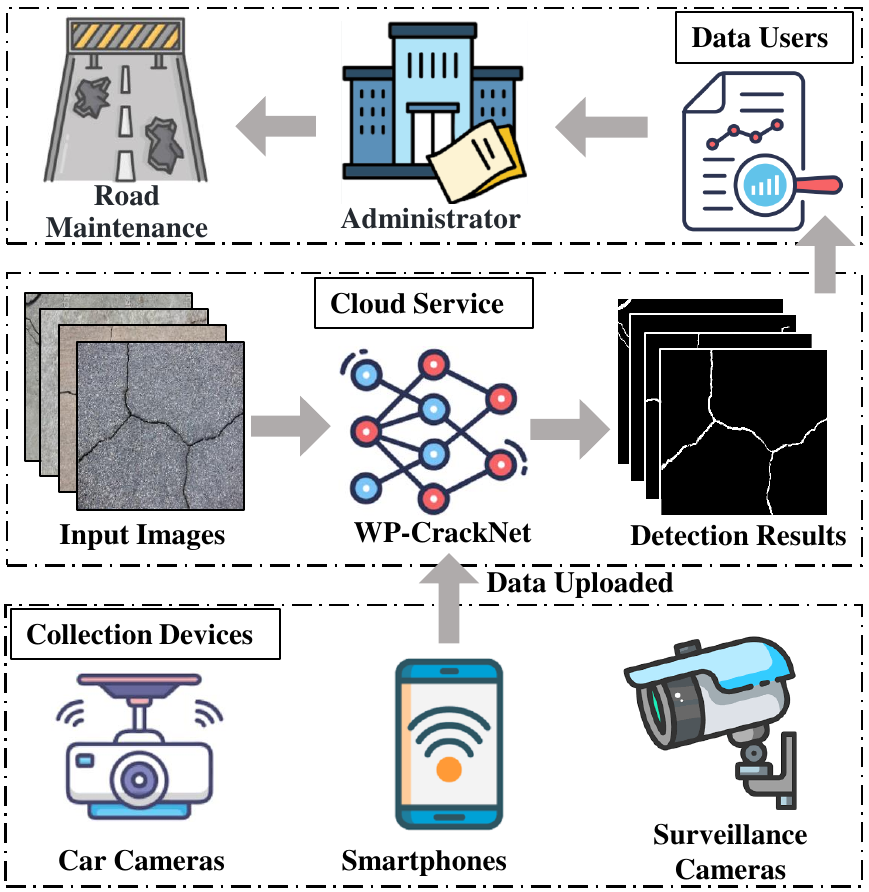}
		\centering
        \captionsetup{font={small}}
		\caption{Road inspection for intelligent infrastructure maintenance with the proposed WP-CrackNet.}
		\label{fig1_IOT_2}
	\end{center}
\end{figure}

Road cracks, typically appearing as narrow lines or curves on road surfaces, serve as early indicators of structural deterioration in civil infrastructure. Although subtle, road cracks can progressively compromise the integrity of roadways and pose significant safety hazards. For instance, in $2020$, deteriorated road conditions contributed to 12.6\% of traffic accidents in the UK \cite{ma2022computer}. Therefore, regular inspection and timely maintenance are crucial to reducing risks and extending the service life of networks \cite{fan2019crack,liu2025these}. At present, manual visual inspection remains the primary approach for crack detection \cite{fan2019road}, but it is costly, labor-intensive, time-consuming, and highly subjective—especially for extensive highway networks exceeding $100,000$ kilometers in countries like China and the US \cite{fan2019pothole}. These limitations underscore the urgent need for automated systems capable of efficiently and objectively analyzing road conditions \cite{fan2021rethinking}. As shown in Fig. \ref{fig1_IOT_2}, collection devices—such as car cameras, smartphones, and surveillance cameras—offer a promising solution by continuously collecting road surface images and uploading the data to cloud-based platforms for automated road crack detection \cite{yuan2020ecrd}. The detection results can provide actionable insights for infrastructure administrators, thereby facilitating more timely and data-driven maintenance decisions.


Recent advancements in deep learning, particularly Convolutional Neural Networks (CNNs) and Transformer-based networks, have significantly enhanced automated road crack detection, which are generally categorized into three types: (1) image classification networks that distinguish between crack and non-crack images \cite{krizhevsky2017imagenet, fan2021deep, hou2021mobilecrack,guo2024lix}; (2) object detection networks that identify cracks at the instance level (location and class) \cite{khan2025real, khan2024real, tsung2025hpph}; and (3) semantic segmentation networks that provide pixel-level crack detection and have emerged as the dominant approach in this field \cite{huyan2020cracku, zhang2023ecsnet, zhu2024lightweight, song2025robust, ma2025self}. However, the training of supervised semantic segmentation networks relies heavily on manually annotated datasets, which are costly and time-consuming to construct. This reliance limits the scalability of road surface perception, as the massive amounts of collected data are difficult to efficiently annotate at the pixel level, thereby hindering the widespread deployment in large-scale infrastructure maintenance.  

To overcome this limitation, we propose a novel \uline{\textbf{W}eakly-supervised \textbf{P}ixel-wise \textbf{C}rack Detection \textbf{Net}work (WP-CrackNet)} via adversarial mutual learning. Unlike prior methods that rely on offline pseudo-label generation \cite{kweon2021unlocking, yoon2022adversarial, kweon2023weakly} or hand-crafted cues \cite{konig2022weakly}, WP-CrackNet adopts an end-to-end training strategy that jointly optimizes three synergistic components—a classifier, a reconstructor, and a detector-using only image-level labels. This design simplifies the training pipeline and promotes a more stable learning process. The classifier generates CAMs to localize discriminative regions, while the reconstructor measures the inferability between road and crack features to enhance structural understanding and encourage crack CAMs to fully cover crack regions (inspired by \cite{kweon2023weakly}). The detector then produces pixel-wise road crack predictions based on pseudo labels derived from post-processed crack CAMs.
Through alternating adversarial training between the classifier and reconstructor, of which outputs feed into the detector, WP-CrackNet achieves stable and mutually reinforcing optimization. Furthermore, we design the PAAM to effectively integrate high-level semantic information with low-level structural cues from the classifier and reconstructor to improve detection performance. Additionally, we introduce the CECCM to refine crack CAMs using center Gaussian weighting and consistency constraints for better generation of pseudo labels. Experimental evaluations on three crack datasets demonstrate that WP-CrackNet not only achieves detection performance comparable to supervised methods but also surpasses SoTA general and road crack detection-specific weakly-supervised methods. The main contributions can be summarized as follows:

\begin{enumerate}
    \item We propose WP-CrackNet, a novel end-to-end weakly-supervised pixel-wise crack detection method using only image-level labels. 
    \item We utilize an adversarial mutual learning strategy for three components which improves learning stability and road crack detection performance.
    \item We design a path-aware attention module to effectively integrate semantic context and structural cues and a center-enhanced CAM consistency module to refine crack CAMs for better generation of pseudo labels.
    \item We validate WP-CrackNet on three self-created datasets, showing comparable performance to supervised methods and outperforming other weakly-supervised methods.  
\end{enumerate}

The remainder of this paper is organized as follows: Section \ref{sec.related} reviews related work. Section \ref{sec.method} details the proposed methodology. Section \ref{sec.experiment} presents implementation details, ablation studies, and comparative experiments. Finally, Section \ref{sec.conclusion} concludes the paper.

\begin{figure*}[!t]
	\begin{center}
		\centering
		\includegraphics[width=0.98\textwidth]{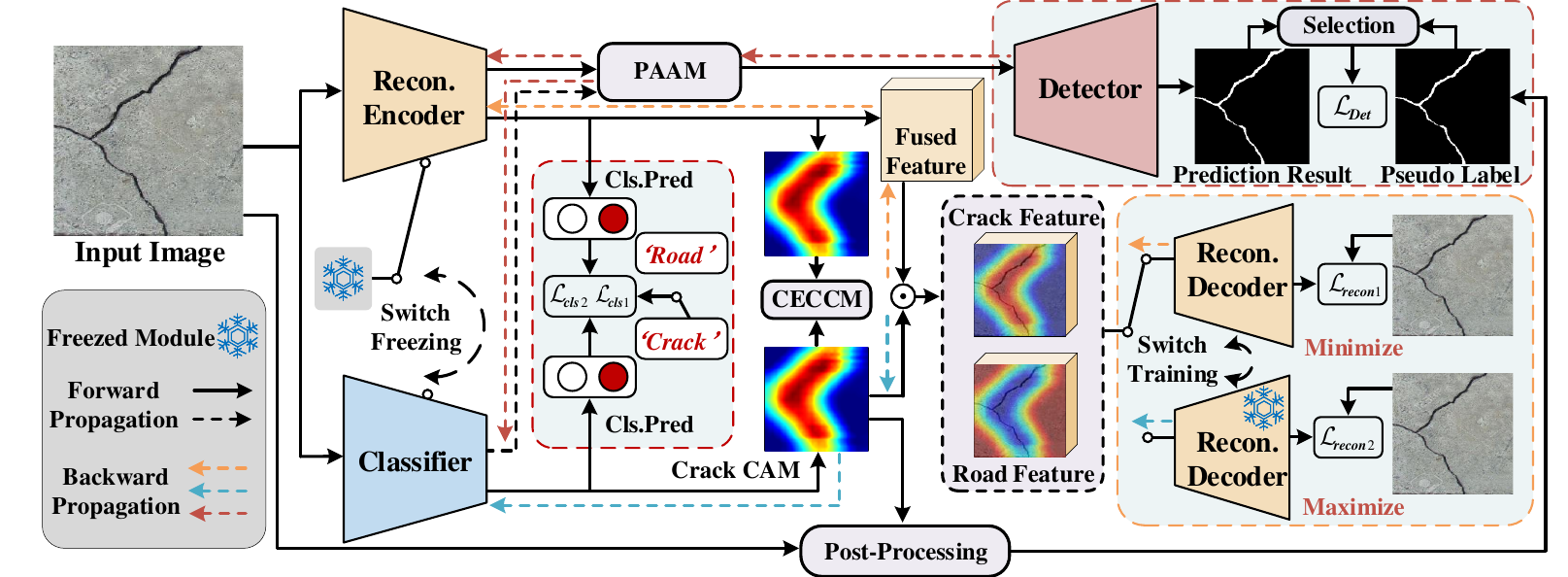}
		\centering
        \captionsetup{font={small}}
		\caption{The overall architecture of WP-CrackNet, consisting of a classifier, a reconstructor, and a detector. During the training phase, the classifier generates CAMs to localize discriminative regions, the reconstructor measures feature inferability, and the detector learns from pseudo labels derived from post-processed crack CAMs to produce pixel-wise road crack detection results.}
		\label{fig1_algorithm}
	\end{center}
\end{figure*}

\section{Related Work}
\label{sec.related}

\subsection{Deep Learning-based Road Crack Detection Methods}
Supervised methods based on CNNs and Transformers have been extensively developed for road crack detection, employing architectures such as FCN \cite{dung2019autonomous}, SegNet \cite{chen2020pavement}, Deeplab \cite{sun2022dma}, U-Net \cite{huyan2020cracku}, \etc. Among them, \cite{zou2018deepcrack} proposed Deepcrack, by fusing features from different scales of SegNet \cite{badrinarayanan2017segnet} to obtain hierarchical details, subsequently resulting in accurate pixel-wise road crack detection results. Similarly,\cite{yang2019feature} proposed a feature pyramid-based hierarchical boosting network (FPHBN), introducing a side network on the HED network \cite{xie2015holistically} to learn hierarchical feature information for road crack detection. Another Deepcrack version \cite{liu2019deepcrack} incorporated a side-output layer into VGG-16 \cite{simonyan2014very}, adopting guided filtering and conditional random field techniques to achieve improved road crack detection performance. \cite{tao2023convolutional} proposed a novel convolutional-transformer network to combine both local and global information extracted from road crack images. In addition, a boundary awareness module was designed to capture boundary details of road cracks to refine crack detection results. Nevertheless, the methods mentioned above are data-driven, and training them relies on massive pixel-level human-annotated labels. The process of obtaining such fine labels is significantly time-consuming and laborious. To reduce this burden, label-efficient methods have been explored. Despite their potential, these methods still face notable limitations. The unsupervised method in \cite{ma2024up} struggles to detect fine cracks due to their subtle visual characteristics. The semi-supervised method in \cite{liu2024semi} remains sensitive to the quality and quantity of labeled data, with performance degrading significantly when labels are scarce. The weakly-supervised method in \cite{konig2022weakly} relies on manually crafted image processing techniques to generate pseudo labels from CAMs offline, leading to unstable outcomes and suboptimal performance.

\subsection{Weakly-supervised Semantic Segmentation Methods}

Weakly-supervised semantic segmentation (WSSS) reduces reliance on costly pixel-level annotations by using weaker supervision such as scribbles \cite{zhang2024scribble}, bounding boxes \cite{khoreva2017simple}, and image-level labels \cite{chang2020weakly, fan2020cian, zhang2021complementary}. Among these, image-level labels are favored for their simplicity and scalability but provide only class presence without spatial information, making precise localization difficult. To tackle this, researchers tend to leverage CAMs \cite{zhou2016learning} to highlight discriminative regions learned by classifiers and use refined CAMs as pseudo labels to train a segmentation network. However, CAMs typically focus on the most distinctive parts, failing to cover the entire target class region. To explicitly expand the CAMs, methods adopting sub-category classification \cite{chang2020weakly}, cross-image relationships \cite{fan2020cian}, contrastive
learning \cite{xie2022c2am}, attention modules \cite{wang2020self} \cite{wu2021embedded} and adversarial erasing mechanisms \cite{kweon2021unlocking, yoon2022adversarial, kweon2023weakly} have been proposed. For instance, \cite{kweon2021unlocking} used a pre-trained classifier to erase discriminative regions, enabling more precise CAM generation by encouraging activation of less discriminative regions. However, the usage of fixed classifier limits its performance. \cite{yoon2022adversarial} further introduced a triplet learning framework that relaxes reliance on the fixed classifier, enabling more flexible guidance of the erasing process and resulting in more complete CAMs. \cite{kweon2023weakly} formulated WSSS as adversarial learning of the classifier and the reconstructor, using the reconstruction task to obtain an effective regularization of CAM generation. Inspired by the methods described above, we proposed WP-CrackNet for end-to-end weakly-supervised road crack detection. Unlike these methods relying on multi-stage processing, our method integrates feature extraction, crack localization, and segmentation into a unified network, further unlocking the potential of mutual adversarial learning. Also, PAAM and CECCM are designed to improve the quality of the generated crack CAMs and enhance the detection performance. The extensive experimental results demonstrate the superiority of WP-CrackNet over SoTA weakly-supervised semantic segmentation methods on the task of road crack detection.

\section{Proposed Methodology}
\label{sec.method}

\subsection{Architecture Overview}

Let the road image training set be $\mathcal{I}^{R}=\{(\boldsymbol{I}_{1}^{R}, \boldsymbol{T}^{R}),..., (\boldsymbol{I}_{n}^{R}, T^{R})\}$ and the crack image training set be 
$\mathcal{I}^{C}=\{(\boldsymbol{I}_{1}^{C}, \boldsymbol{T}^{C}),...,(\boldsymbol{I}_{m}^{C}, \boldsymbol{T}_{m}^{C})\}$, where $\boldsymbol{T}^{R}=[1,0]$ denotes the non-crack class label and $\boldsymbol{T}^{C}=[0,1]$ denotes the crack class label, respectively. Here, $\boldsymbol{I}_{i}^{R}$, $\boldsymbol{I}_{i}^{C} \in \mathbb{R}^{H \times W \times 3}$ denote the $i$-th road and crack image with $H$ and $W$ as the height and width.
The overall architecture of WP-CrackNet is illustrated in Fig. \ref{fig1_algorithm}, comprising a classifier $Cls$, a reconstructor $Rec$, and a detector $Det$. The classifier is trained via image-level supervision and generates crack CAMs to highlight crack regions within the crack image. The reconstructor is designed with an encoder-decoder architecture, aiming to assess the inferability of crack versus non-crack regions by reconstructing the input crack image. The detector is trained using pseudo labels derived from the post-processed crack CAMs to output refined pixel-wise crack detection results.

During the training phase, multi-layer fused feature maps from the encoder of reconstructor $Rec.E$ are decomposed into crack and road features based on crack CAMs, which are reconstructed separately. Inspired by \cite{kweon2023weakly}, when crack CAMs fully cover all crack regions, the inferability between crack and road features becomes low. To leverage this, an adversarial scheme trains $Rec$ to reconstruct one feature from the other, while $Cls$ learns to generate crack CAMs that hinder this reconstruction. Simultaneously, $Det$ refines its predictions by combining high-level semantic information from $Cls$ with low-level structural cues from $Rec$, guided by the post-processed crack CAMs. This collaborative framework forms a reciprocal feedback loop among the three modules, improving training stability and detection performance under weak supervision.

\subsection{Adversarial training of classifier and reconstructor}

Given an input image $\boldsymbol{I} \in \{\boldsymbol{I}^{R}$, $\boldsymbol{I}^{C}\}$, the classifier and reconstructor output class predictions $q$ and $q^{rec}$ indicating the presence of cracks, along with CAMs $\boldsymbol{M}$ and $\boldsymbol{M}^{rec}$ that highlight discriminative regions. This process is formulated as:
\begin{equation}
    \boldsymbol{M},\ q = Cls(\boldsymbol{I}), \quad
    \boldsymbol{M}^{rec},\ q^{rec} = Rec.E(\boldsymbol{I}).
\end{equation}
In line with \cite{zhang2018adversarial}, ResNet38 \cite{wu2019wider} is employed as the backbone network for both $Cls$ and $Rec.E$, followed by a $1\times1$ convolution layer serving as the classification head to generate CAMs.
For input crack image $\boldsymbol{I}^{C}$, multi-layer fused feature map $\boldsymbol{Z}$ is obtained by passing it through $Rec.E$, and is decomposed into crack feature map $\boldsymbol{Z}_C$ and road feature map $\boldsymbol{Z}_R$ by using the corresponding crack CAM $\boldsymbol{M}_{C}$:

\begin{equation}
    \boldsymbol{Z}_{C} = \boldsymbol{Z} \odot \boldsymbol{M}_{C}, \quad
    \boldsymbol{Z}_{R} = \boldsymbol{Z} \odot (1 - \boldsymbol{M}_{C}),
\end{equation}

where $\odot$ denotes element-wise multiplication. Then, a switch training strategy is adopted, where $Cls$ and $Rec.E$ are updated in turn. $\boldsymbol{Z}_C$ and $\boldsymbol{Z}_R$ are passed into the decoder of reconstructor $Rec.D$ (using a UNet-based network) to obtain corresponding reconstruction results:

\begin{equation}
    \boldsymbol{O}_C = Rec.D(\boldsymbol{Z}_{C}), \quad
    \boldsymbol{O}_{R} = Rec.D(\boldsymbol{Z}_{R}). 
\end{equation}
When $Cls$ is frozen, $Rec.E$ is trained to reconstruct one feature from the other. Conversely, with $Rec.E$ frozen, $\hat{\boldsymbol{O}}_C$ and $\hat{\boldsymbol{O}}_R$ are obtained, and $Cls$ is trained to generate $\boldsymbol{M}_{C}$ that hinders the reconstruction of the original image.

Furthermore, considering the potential activation drift or instability in CAMs introduced by the adopted adversarial training scheme, along with the inherently slender and low-contrast nature of cracks, we propose a center-enhanced CAM consistency module (CECCM) to better guide the generation of $\boldsymbol{M}_{C}$. This module enhances spatial alignment between $Cls$ and $Rec.E$ by applying Gaussian-based center weighting and enforcing consistency between the center-enhanced crack CAMs. Given an input crack image, $\boldsymbol{M}_{C}$ and $\boldsymbol{M}^{rec}_{C}$ are obtained from $Cls$ and $Rec.E$, respectively. Then, a center-enhancement operation is applied to both CAMs, guided by their spatial center of mass. Specifically, for a CAM $M$, we compute its normalized spatial center as:
\begin{equation}
\mu_x = \frac{\sum_{x,y} x \cdot \boldsymbol{M}(x,y)}{\sum_{x,y} \boldsymbol{M}(x,y)}, \quad
\mu_y = \frac{\sum_{x,y} y \cdot \boldsymbol{M}(x,y)}{\sum_{x,y} \boldsymbol{M}(x,y)}.
\end{equation}

Then, a spatial Gaussian prior $\boldsymbol{G}\in \mathbb{R}^{H \times W}$ centered at $(\mu_x, \mu_y)$ is constructed:

\begin{equation}
\boldsymbol{G}(x, y) = \exp\left(-\frac{(x - \mu_x)^2 + (y - \mu_y)^2}{2\sigma^2}\right),
\end{equation}
where $\sigma$ controls the spread of the Gaussian. Then, the center-enhanced CAMs are obtained via element-wise multiplication:
\begin{equation}
\boldsymbol{M}_{CG} = \boldsymbol{M}_{C} \odot \boldsymbol{G}_{C},
\quad
\boldsymbol{M}^{rec}_{CG} = \boldsymbol{M}^{rec}_{C} \odot \boldsymbol{G}^{rec}_{C}.
\end{equation}
Finally, we impose a consistency loss between the two center-enhanced CAMs to encourage $Cls$ and $Rec.E$ to produce spatially aligned and structure-consistent activations.


\begin{figure}[!t]
	\begin{center}
		\centering
		\includegraphics[width=0.49\textwidth]{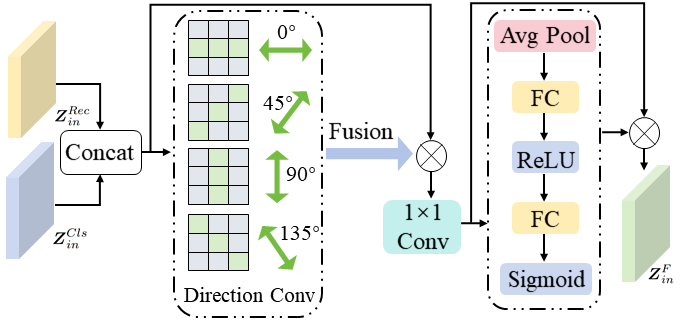}
		\centering
        \captionsetup{font={small}}
		\caption{The proposed PAAM, consisting of a spatial attention branch and a channel attention branch.}
		\label{fig3}
	\end{center}
\end{figure}

\subsection{Iterative Training of Detector}
\label{inter}

Given an input crack image $\boldsymbol{I}^{C}$, we empirically select the output feature map from an intermediate residual block in both $Cls$ and $Rec.E$ as the input to $Det$. This layer leverages dilated convolutions to extract abundant semantic representations while maintaining a relatively high spatial resolution. The obtained feature maps are denoted as $\boldsymbol{Z}^{Cls}_{in}$ and $\boldsymbol{Z}^{Rec}_{in}$, which provide high-level semantic information and low-level structural cues for the training of $Det$.

To effectively fuse these features, we concatenate them as $\boldsymbol{Z}^{Con}_{in}$ and feed the result into a novel attention mechanism named Path-Aware Attention Module (PAAM), which enhances the discriminative capability of the fused representation, especially by adaptively emphasizing crack-relevant regions through modeling both spatial and channel-wise dependencies. The structure of PAAM is illustrated in Fig. \ref{fig3}, consisting of a spatial attention branch and a channel attention branch. To capture the directional characteristics of cracks,
we apply directional convolutions $D_{\theta}(\cdot)$ along four orientations $\theta \in \{0^{\circ},45^{\circ},90^{\circ},135^{\circ}\}$. For each direction, we compute the absolute response: 
\begin{equation}
    R_{\theta} = | D_{\theta}(\boldsymbol{Z}^{Con}_{in}) |,
\end{equation}
and the spatial attention map $A_{spatial}$ is obtained by aggregating the directional responses followed by a sigmoid activation $\delta$:
\begin{equation}
    A_{spatial} = \delta(R_{0^{\circ}}+R_{45^{\circ}}+R_{90^{\circ}}+R_{135^{\circ}}).
\end{equation}
The spatial attention map serves as a soft mask that highlights potential crack paths across the feature map. We perform path-weighted fusion by applying this attention to the input feature via element-wise multiplication, followed by a convolution block with $1\times1$ kernel size to reduce computational cost and refine the fused features:
\begin{equation}
\boldsymbol{Z}^{spatial}_{in} = {ReLU}({BN}(W_{1\times1} (\boldsymbol{Z}^{Con}_{in} \odot A_{spatial}))).
\end{equation}

Then, $\boldsymbol{Z}^{spatial}_{in}$ is fed into the channel attention branch, which models inter-channel dependencies by emphasizing informative feature channels to better capture complementary cues such as texture, edges, and context for crack detection. Following the squeeze-and-excitation paradigm, we apply a global context aggregation using adaptive average pooling to obtain a channel-wise descriptor:

\begin{equation}
\boldsymbol{s} = AvgPool(\boldsymbol{Z}^{spatial}_{in}) \in \mathbb{R}^{C \times 1 \times 1}.
\end{equation}
The descriptor is then reshaped into a vector and passed through two fully connected layers, interleaved with ReLU and sigmoid activations, to generate the channel attention map:

\begin{equation}
    A_{channel} = \delta(Fc(ReLU(Fc(s_{flatten})))).
\end{equation}
Finally, the reshaped channel attention map is element-wise multiplied with $\boldsymbol{Z}^{spatial}_{in}$ to produce $\boldsymbol{Z}^{F}_{in}$, which serves as the input to $Det$ (composed of a series of convolution and transposed convolution layers). 

To enable the network to learn the mapping from input crack image to pixel-wise road crack detection results $\boldsymbol{Y}_{out}$, we train $Det$ using the pseudo label $\boldsymbol{Y}_{pse}$, which is obtained by post-processing the corresponding crack CAM with dense Conditional Random Fields (denseCRF) \cite{krahenbuhl2011efficient}. The denseCRF algorithm refines the coarse CAM by modeling long-range dependencies between pixels based on both spatial proximity and color similarity. It encourages pixels with similar appearances and close spatial distances to be assigned the same label, which is particularly effective for road crack detection where cracks are often thin and low-contrast. 

Since the quality of crack CAMs evolves during training, the pseudo labels are dynamically updated, making the learning process inherently iterative. To further improve training robustness, we introduce a selection mechanism that filters out pseudo labels that are entirely empty (\emph{i.e., all-black masks}), thereby avoiding supervision from uninformative samples and enhancing the quality of the learning process.


\subsection{Loss Function}

For the training of WP-CrackNet, the total loss function is defined as follows:
\begin{equation}
    \mathcal{L}_{total} = \mathcal{L}_{Rec} + \mathcal{L}_{Cls}+\mathcal{L}_{Det},
\end{equation}
where $\mathcal{L}_{Rec}$, $\mathcal{L}_{Cls}$, and $\mathcal{L}_{Det}$ denote the training losses for $Rec$, $Cls$ and $Det$. During training, $\mathcal{L}_{Rec}$ and $\mathcal{L}_{Cls}$ are alternately optimized with the other module frozen, and the parameters of $Det$ are consistently updated with $\mathcal{L}_{Det}$ to guide precise road crack detection.
 
\subsubsection{Reconstructor Loss}
Given an input image $\boldsymbol{I}$, we first use the standard binary cross-entropy (BCE) loss between the class prediction $q^{rec}$ and image-level ground-truth label $\boldsymbol{T} \in \{\boldsymbol{T}^{R},\boldsymbol{T}^{C}\}$ to facilitate the learning of classification ability, denoted as:
\begin{equation}
    \mathcal{L}_{cls1} = - \left[ \boldsymbol{T} \log \delta(q^{rec}) + (1 - \boldsymbol{T}) \log(1 - \delta(q^{rec})) \right].
\end{equation}
As mentioned above, we train $Rec$ to reconstruct one feature from the other. To ensure consistency, we minimize the difference between the reconstructed result and the input crack image within the opposite class region. Specifically, for crack and road features, we minimize the following losses:
\begin{equation}
    \mathcal{L}^{C}_{recon1} = |(1-\boldsymbol{M}^{rec}_{C})  \odot (\boldsymbol{I}_{C}-\boldsymbol{O}_{C}) |_1, 
\end{equation}
\begin{equation}
    \mathcal{L}^{R}_{recon1} = |\boldsymbol{M}^{rec}_{C}  \odot (\boldsymbol{I}_{C}-\boldsymbol{O}_{R}) |_1,
\end{equation}
where $|\cdot|_1$ represents L1 loss. Therefore, the final reconstructor loss is formulated as follows:
\begin{equation}
    \mathcal{L}_{Rec} = \lambda_{c1}\mathcal{L}_{cls1} + \lambda^{C}_{r1}\mathcal{L}^{C}_{recon1} + \lambda^{R}_{r1}\mathcal{L}^{R}_{recon1},
\end{equation}
where $\lambda_{c1}$, $\lambda^{C}_{r1}$, and $\lambda^{R}_{r1}$ are weighting parameters used to harmonize these losses, and we have $\mathcal{L}_{recon1}=\lambda^{C}_{r1}\mathcal{L}^{C}_{recon1} + \lambda^{R}_{r1}\mathcal{L}^{R}_{recon1}$.

\subsubsection{Classifier Loss}
Given an input image $\boldsymbol{I}$, similarly, the BCE loss is employed to facilitate the learning of classification ability for $Cls$:
\begin{equation}
    \mathcal{L}_{cls2} = - \left[ \boldsymbol{T} \log \delta(q) + (1 - \boldsymbol{T}) \log(1 - \delta(q)) \right].
\end{equation}
For input crack images, the objective of $Cls$ is to generate crack CAMs that hinder $Rec$ from reconstructing the original image. To this end, we minimize the similarity between the reconstruction results and the input image on crack region, with a similar constraint also applied to the road region:
\begin{equation}
    \mathcal{L}^{C}_{recon2} = -|(1-\boldsymbol{M}_{C})  \odot (\boldsymbol{I}_{C}-\hat{\boldsymbol{O}}_{C}) |_1,
\end{equation}
\begin{equation}
    \mathcal{L}^{R}_{recon2} = -|\boldsymbol{M}_{C}  \odot (\boldsymbol{I}_{C}- \hat{\boldsymbol{O}_{R}})|_1.
\end{equation}
Furthermore, a consistency loss between center-enhanced crack CAMs from $Cls$ and $Rec$ is designed to enforce spatial alignment and enhance the focus on central crack regions, thereby facilitating more accurate and robust crack localization, which is denoted as:
\begin{equation}
    \mathcal{L}_{CEC} = \frac{1}{HW} \sum |\boldsymbol{M}_{CG}-  \boldsymbol{M}^{rec}_{CG}|.
\end{equation}
Therefore, the final classifier loss is formulated as follows:
\begin{align}
    \mathcal{L}_{Cls} =\ & \lambda_{c2} \mathcal{L}_{cls2} + \lambda^{C}_{r2} \mathcal{L}^{C}_{recon2} \nonumber \\
    & + \lambda^{R}_{r2} \mathcal{L}^{R}_{recon2} + \lambda_{cec} \mathcal{L}_{CEC},
\end{align}
where $\lambda_{c2}$, $\lambda^{C}_{r2}$, $\lambda^{R}_{r2}$, and $\lambda_{cec}$ are weighting parameters used to harmonize these losses. We have $\mathcal{L}_{recon2}=-\lambda^{C}_{r2} \mathcal{L}^{C}_{recon2} - \lambda^{R}_{r2} \mathcal{L}^{R}_{recon2}$.

\subsubsection{Detector Loss}
Given input crack image ${I}_{C}$, as illustrated in \ref{inter}, we can obtain road crack detection results $\boldsymbol{Y}_{out}$ by passing through $Cls$, $Rec$, and $Det$. The detector is trained by using the corresponding pseudo label $\boldsymbol{Y}_{pse}$, and the detector loss is denoted as:
\begin{align}
\mathcal{L}_{Det} = - \frac{1}{HW} \sum \Big[ 
& \boldsymbol{Y}_{pse} \log \delta(\boldsymbol{Y}_{out}) \nonumber+ \\
&  (1 - \boldsymbol{Y}_{pse}) \log \left(1 - \delta(\boldsymbol{Y}_{out}) \right) 
\Big].
\end{align}

\section{Experimental Results}

\label{sec.experiment}

\begin{table*}[t!]
	\renewcommand{\arraystretch}{1}
        \captionsetup{font={small}}
	\caption{
Ablation study results for pixel-wise crack detection performance to investigate the impact of integrating outputs of $Cls$ and $Rec.E$, and to validate the effectiveness of the proposed CECCM and PAAM on the DeepCrack dataset \cite{liu2019deepcrack}. The symbol \Checkmark indicates the used module for the training of the detection branch.
 }
	\centering
	\begin{tabular}{R{0.7cm}R{0.7cm}R{0.7cm}R{1.1cm}R{1.1cm}R{1.7cm}R{1.9cm}R{1.8cm}R{2.2cm}R{1.6cm}}
		\toprule
		{$Cls$}&{$Det$}&{$Rec.E$} & CECCM & PAAM & Precision~($\%$)$\uparrow$ & Recall~($\%$)$\uparrow$ & Accuracy~($\%$)$\uparrow$ & F1-Score~($\%$)$\uparrow$ & IoU~($\%$)$\uparrow$ \\ \midrule
		{\Checkmark}&{\Checkmark}&{}&{}&{} & 90.111 & 68.308 & 98.305 &  77.709 & 63.545 \\
        {\Checkmark}&{\Checkmark}&{\Checkmark}&{}&{} & 83.920 & 74.032 & 98.263 & 78.666 & 64.835 \\
        {\Checkmark}&{\Checkmark}&{\Checkmark}&{\Checkmark}&{} & 83.828 & 78.341 & 98.409 & 80.992 & 68.055 \\
        {\Checkmark}&{\Checkmark}&{\Checkmark}&{\Checkmark}&{\Checkmark} & 82.868 & 80.682 & 98.443 & \textbf{81.760} & 
\textbf{69.148} \\
		\bottomrule
		\\
	\end{tabular}
	\label{table1}
\end{table*}

\subsection{Datasets}

The \textbf{Crack500} \cite{yang2019feature} dataset consists of $500$ high-resolution road images ($2000\times1500$ pixels) with pixel-level annotations. Each image is divided into $16$ non-overlapping regions, and those with over $1,000$ crack pixels are retained, yielding $1,896$ training, $348$ validation, and $1,124$ test images. This dataset includes four crack types and poses challenges such as shadows, occlusions, and varying lighting. For our experiments, supervised methods are trained on the original dataset with all images resized to $256\times256$ pixels for consistency. Additionally, we create a new training set by combining $756$ undamaged road images cropped from the original dataset with $1,896$ crack images, and assign image-level labels to facilitate the training of weakly-supervised methods.

The \textbf{DeepCrack} \cite{liu2019deepcrack} dataset consists of $537$
pixel-level annotated images of cracks on concrete and asphalt surfaces across various scenes and scales. Each image is captured at a resolution of $544\times384$ pixels and split into $300$ training and 
$237$ test images. Similarly, we train supervised methods on the original dataset with the size of $256\times256$ pixels and create a new training set for weakly-supervised methods by combining $296$ undamaged road images (cropped and rescaled from the original dataset) with $300$ crack images, all labeled at the image level.

The \textbf{CFD} \cite{shi2016automatic} dataset consists of $118$ high-quality images of concrete surfaces with cracks ranging from $1$ to $3$ $mm$ in width, each annotated at the pixel level with a resolution of $480\times320$ pixels. The dataset features diverse illumination conditions, increasing the difficulty of accurate crack detection. For experiments, $70$ images are used for training and $48$ for testing, all resized to $256\times256$ for the training of supervised methods. To support weakly-supervised methods and account for the subtle nature of cracks in this dataset, we enlarge each image in the training set and divide it into a $3\times3$ grid of patches. This results in a new training set consisting of $268$ crack images and $256$ undamaged images, all annotated with image-level labels.

\subsection{Implementation Details}

All experiments are conducted on a single NVIDIA RTX 4090 GPU, with each model trained for $200$ epochs. The initial learning rate is set to $0.001$ and adjusted dynamically using the polynomial decay policy. To enhance generalization, standard data augmentation techniques such as random cropping, resizing, and horizontal flipping are applied to the input images. Following the empirical settings suggested in \cite{kweon2023weakly}, we configure the loss weights as follows: for $\mathcal{L}_{Rec}$, we set $\lambda_{c1}=1$ and $\lambda^{C}_{r1}=\lambda^{R}_{r1}=0.5$; for $\mathcal{L}_{Cls}$, we use $\lambda_{c2}=1$, $\lambda^{C}_{r2}=0.8$, $\lambda^{R}_{r2}=0.3$, and $\lambda_{cec}=0.5$.

During inference on the CFD dataset, each image is divided into a
$3\times3$ grid of patches, which are resized and classified by $Cls$ to filter out background regions. Only patches predicted to contain cracks are passed to the detection module, and final result is obtained by merging the outputs from these selected patches. For fair comparison, all weakly-supervised methods for comparison are processed using denseCRF refinement following CAM generation. Additionally, they are trained using the same $Rec$ and $Det$ architecture as our method. For evaluation, we adopt a comprehensive set of metrics, including precision, recall, accuracy, IoU, and F1-score, to quantitatively assess the detection performance of WP-CrackNet against existing methods. In addition, model parameters and frames per second (FPS) are used to evaluate the model complexity and processing speed.

\subsection{Ablation Study}

To assess the impact of integrating both high-level semantic information and low-level structural cues from $Cls$ and $Rec.E$, and to evaluate the effectiveness of the proposed CECCM and PAAM, we conduct an ablation study on the DeepCrack dataset \cite{liu2019deepcrack}. The quantitative experimental results are presented in Table \ref{table1}.
The first row shows crack detection results using only $Cls$ and $Det$, while the last row represents the process of obtaining fused feature map through $Cls$ and $Det$, which is then enhanced by PAAM, with crack CAMs further improved by the CECCM for better generation of pseudo labels. The experimental results that integrate all these modules attain the best detection performance, demonstrating the effectiveness of the proposed CECCM and PAAM. 

\begin{figure}[t!]
	\begin{center}
	\includegraphics[width=0.49\textwidth]{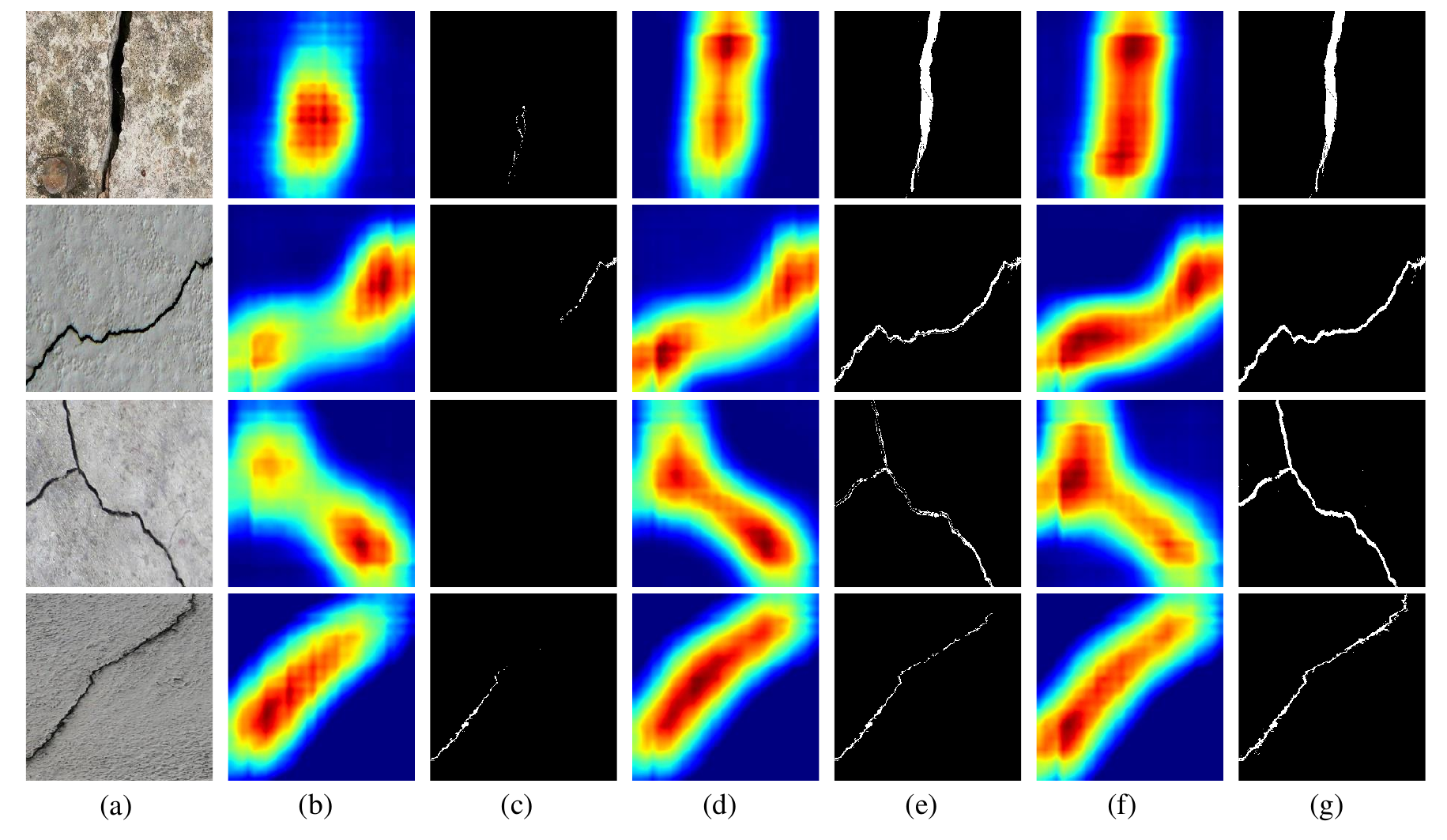}
        \captionsetup{font={small}}
	\caption{Ablation study results illustrating crack CAMs (b, d, f) obtained under different training strategies — classifier only, adversarial training, and adversarial training with CECCM — and their corresponding pseudo labels (c, e, g) produced via denseCRF, with (a) showing the input image.}
		\label{fig_ablation2}
	\end{center}
\end{figure}

In addition, qualitative experiments are conducted to visually validate the effectiveness of the adopted adversarial training strategy and the proposed CECCM. The results in Fig. \ref{fig_ablation2} indicate that the adversarial training strategy enables crack CAMs to fully cover crack regions, while the proposed CECCM enhances the generation and spatial aggregation of crack CAMs. Furthermore, with the aid of denseCRF, high-quality pseudo labels can be derived for the training of pixel-wise road crack detection. In Fig. \ref{fig_ablation2}, (a) is the input image; (b), (d), and (f) are crack CAMs obtained under different training settings (classifier only, adversarial training, and adversarial training with CECCM, respectively); and (c), (e), and (g) are their corresponding pseudo labels after applying denseCRF.

\begin{table*}[t!]
	\renewcommand{\arraystretch}{1}
        \captionsetup{font={normalsize}}
	\caption{Quantitative experimental results of pixel-wise crack detection performance on the Crack500 dataset \cite{yang2019feature}.}
	\centering
    \begin{threeparttable}
	\begin{tabular}{C{2.2cm}R{2.7cm}R{1.7cm}R{1.7cm}R{2.1cm}R{2.2cm}R{1.6cm}}
		\toprule
		\multicolumn{1}{r}{Training Strategy}&{Methods} & Precision~($\%$)$\uparrow$ & Recall~($\%$)$\uparrow$ & Accuracy~($\%$)$\uparrow$ & F1-Score~($\%$)$\uparrow$ & IoU~($\%$)$\uparrow$ \\ \midrule
		\multirow{5}{*}{\raisebox{0pt}[0pt][0pt]{\makecell{Specific Supervised}}}
        &{Deepcrack19 \cite{liu2019deepcrack}} & 57.581 & 86.733 & 95.687 & 69.213 & 52.920 \\
        &{Deepcrack18 \cite{zou2018deepcrack}} & 70.919 & 70.356 & 96.731 & 70.636 & 54.603 \\
        &{Crack-Att \cite{xu2023crack}} & 66.497 & 70.883 & 96.376 & 68.620 & 52.230 \\
        &{HrSegNet \cite{li2023real}}  & 67.652 & 74.092 & 96.572 & \textbf{70.726}  & \textbf{54.710} \\
        &{CT-crackseg \cite{tao2023convolutional}} & 62.158 & 72.126 & 96.472 & 66.772 & 50.119 \\
        \toprule
        \multirow{6}{*}{\raisebox{0pt}[0pt][0pt]{\makecell{Weakly-Supervised}}}
        &{AEFT \cite{yoon2022adversarial}} & 68.406 & 26.986 & 95.222 & 38.703 & 23.995 \\
	&{OC-CSE \cite{kweon2021unlocking}} & 62.014 & 26.897 & 94.993 & 37.520 & 23.092 \\
        &{VWL \cite{ru2022weakly}} & 85.821 & 4.019 & 94.598 & 7.6782 & 3.992 \\
        &{ACR \cite{kweon2023weakly}}  & 6.580 & 75.302 & 38.859 & 12.103 & 6.441 \\
        &{WS-SCS \cite{konig2022weakly}}  & 72.904 & 38.411 & 95.759 & 50.314 & 33.613 \\
        &{WP-CrackNet} & 71.812 & 55.579 & 96.298 & \textbf{62.661} & \textbf{45.625} \\
		\toprule
		\multirow{1}{*}{\makecell{ Unsupervised}}&{UP-CrackNet \cite{ma2024up}} & 65.377 & 58.609 & 95.484 & 61.808 & 44.726 \\
		\bottomrule
        \\
	\end{tabular}
    \end{threeparttable}
	\label{table_c1}
\end{table*}

\begin{table*}[t!]
	\renewcommand{\arraystretch}{1}
        \captionsetup{font={normalsize}}
	\caption{Quantitative experimental results of pixel-wise crack detection performance on the DeepCrack dataset \cite{liu2019deepcrack}.}
	\centering
    \begin{threeparttable}
	\begin{tabular}{C{2.2cm}R{2.7cm}R{1.7cm}R{1.7cm}R{2.1cm}R{2.2cm}R{1.6cm}}
		\toprule
		\multicolumn{1}{r}{Training Strategy}&{Methods} & Precision~($\%$)$\uparrow$ & Recall~($\%$)$\uparrow$ & Accuracy~($\%$)$\uparrow$ & F1-Score~($\%$)$\uparrow$ & IoU~($\%$)$\uparrow$ \\ \midrule
		\multirow{5}{*}{\raisebox{0pt}[0pt][0pt]{\makecell{Specific Supervised}}}
        &{Deepcrack19 \cite{liu2019deepcrack}} & 88.785 & 68.923 & 97.756 & 77.603 & 63.403 \\
        &{Deepcrack18 \cite{zou2018deepcrack}} & 71.720 & 88.403 & 98.350 & 79.192 & 65.552 \\
        &{Crack-Att \cite{xu2023crack}} & 89.967 & 69.327 & 98.339 & 78.310 & 64.352 \\
        &{HrSegNet \cite{li2023real}}  & 82.492 & 80.337 & 98.412 & 81.400  & 68.634 \\
        &{CT-crackseg \cite{tao2023convolutional}} & 85.381 & 78.838 & 98.501 & 
        \textbf{81.979} & \textbf{69.461} \\
        \toprule
        \multirow{6}{*}{\raisebox{0pt}[0pt][0pt]{\makecell{Weakly-Supervised}}}
        &{AEFT \cite{yoon2022adversarial}} & 81.899 & 68.112 & 97.969 & 74.372 & 59.199 \\
	&{OC-CSE \cite{kweon2021unlocking}} & 89.786 & 66.131 & 98.209 & 76.164 & 61.504 \\
        &{VWL \cite{ru2022weakly}} & 86.356 & 56.650 & 97.738 & 68.418 & 51.996 \\
        &{ACR \cite{kweon2023weakly}}  & 88.339 & 66.424 & 98.168 & 75.829 & 61.069 \\
        &{WS-SCS \cite{konig2022weakly}}  & 98.499 & 30.722 & 96.952 & 46.836 & 30.579 \\
        &{WP-CrackNet} & 82.868 & 80.682 & 98.443 & \textbf{81.760} & \textbf{69.148} \\
		\toprule
		\multirow{1}{*}{\makecell{ Unsupervised}}&{UP-CrackNet \cite{ma2024up}} & 63.412 & 88.852 & 98.049 & 74.006 & 58.738 \\
		\bottomrule
        \\
	\end{tabular}
    \end{threeparttable}
	\label{table_c2}
\end{table*}

\begin{table*}[t!]
	\renewcommand{\arraystretch}{1}
        \captionsetup{font={normalsize}}
	\caption{Quantitative experimental results of pixel-wise crack detection performance on the CFD dataset \cite{shi2016automatic}.}
	\centering
    \begin{threeparttable}
	\begin{tabular}{C{2.2cm}R{2.7cm}R{1.7cm}R{1.7cm}R{2.1cm}R{2.2cm}R{1.6cm}}
		\toprule
		\multicolumn{1}{r}{Training Strategy}&{Methods} & Precision~($\%$)$\uparrow$ & Recall~($\%$)$\uparrow$ & Accuracy~($\%$)$\uparrow$ & F1-Score~($\%$)$\uparrow$ & IoU~($\%$)$\uparrow$ \\ \midrule
		\multirow{5}{*}{\raisebox{0pt}[0pt][0pt]{\makecell{Specific Supervised}}}
        &{Deepcrack19 \cite{liu2019deepcrack}} & 20.892 & 88.142 & 94.372 & 33.778 & 20.321 \\
        &{Deepcrack18 \cite{zou2018deepcrack}} & 46.231 & 69.159 & 98.188 & 55.417 & 38.329 \\
        &{Crack-Att \cite{xu2023crack}} & 70.086 & 43.530 & 98.778 & 53.705 & 36.710 \\
        &{HrSegNet \cite{li2023real}}  & 29.782 & 43.474 & 98.322 & 35.348  & 21.469 \\
        &{CT-crackseg \cite{tao2023convolutional}} & 60.907 & 54.916 & 98.692 & 
        \textbf{57.757} & \textbf{40.604} \\
        \toprule
        \multirow{6}{*}{\raisebox{0pt}[0pt][0pt]{\makecell{Weakly-Supervised}}}
        &{AEFT \cite{yoon2022adversarial}} & 93.233 & 4.008 & 98.432 & 7.685 & 3.996 \\
	&{OC-CSE \cite{kweon2021unlocking}} & 81.154 & 6.397 & 98.452 & 11.860  & 6.304 \\
        &{VWL \cite{ru2022weakly}} & 94.161 & 3.746 & 98.429 & 7.206  & 3.738 \\
        &{ACR \cite{kweon2023weakly}}  & 80.268 & 4.217 & 98.423 & 8.013 & 4.174 \\
        &{WS-SCS \cite{konig2022weakly}}  & 77.764 & 2.485 & 98.400 & 4.816 & 2.468 \\
        &{WP-CrackNet} & 62.262 & 49.682 & 98.690 & \textbf{55.265} & \textbf{38.184} \\
		\toprule
		\multirow{1}{*}{\makecell{ Unsupervised}}&{UP-CrackNet \cite{ma2024up}} & 10.978 & 63.785 & 90.987 & 18.731 & 10.333 \\
		\bottomrule
        \\
	\end{tabular}
    \end{threeparttable}
	\label{table_c3}
\end{table*}
 
\begin{figure*}[t!]
	\begin{center}
	\centering
	\includegraphics[width=0.95\textwidth]{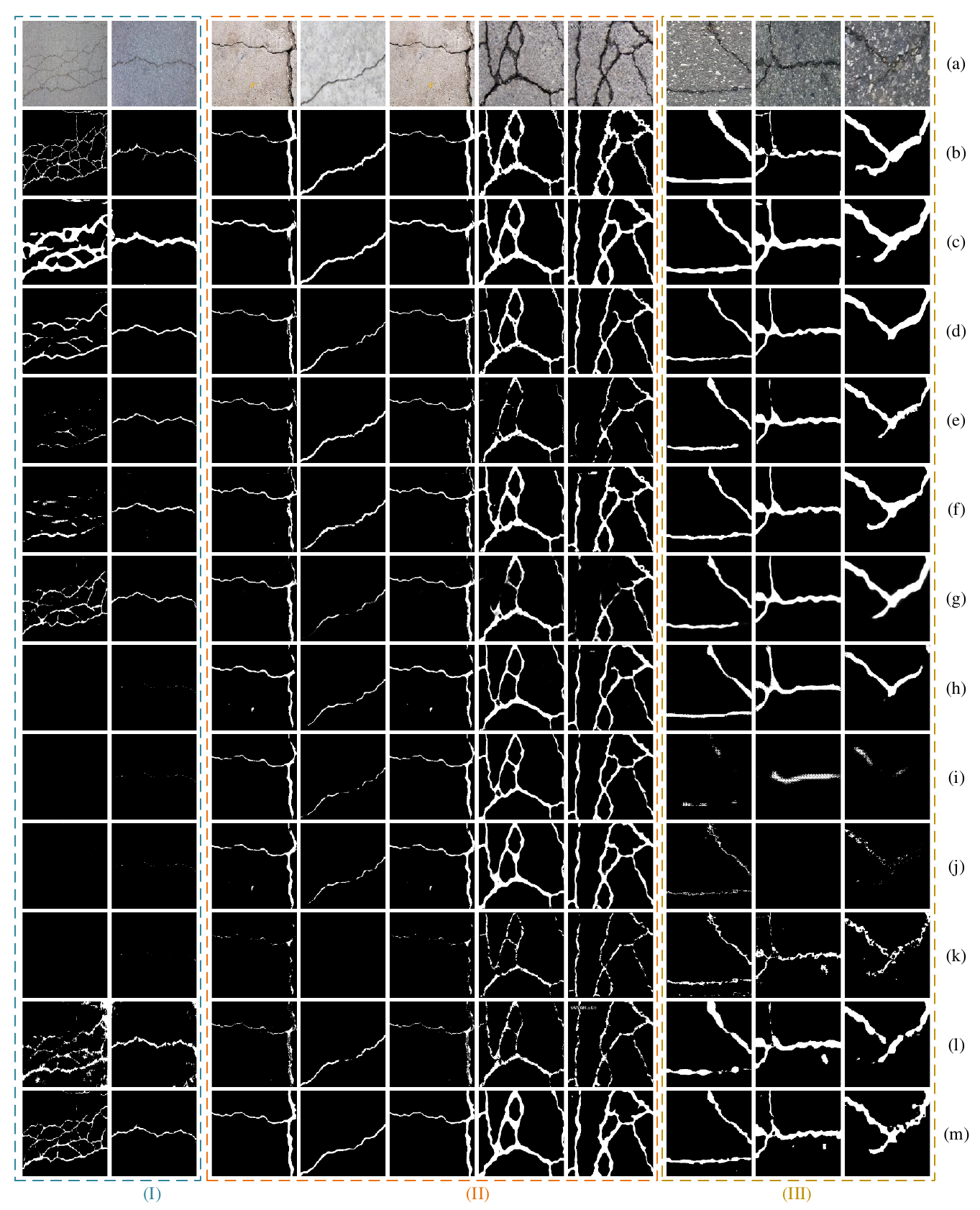}
	\centering
        \captionsetup{font={small}}
	\caption{Examples of experimental results on the (I) CFD \cite{shi2016automatic}, (II) DeepCrack \cite{liu2019deepcrack}, and (III) Crack500 \cite{yang2019feature} datasets: (a) Input images; (b) Ground Truth; (c) Deepcrack19 \cite{liu2019deepcrack}; (d) Deepcrack18 \cite{zou2018deepcrack}; (e) Crack-Att \cite{xu2023crack}; (f) HrSegNet \cite{li2023real}; (g) CT-crackseg \cite{tao2023convolutional}; (h) AEFT \cite{yoon2022adversarial}; (i) OC-CSE \cite{kweon2021unlocking}; (j) ACR \cite{kweon2023weakly}; (k) WS-SCS \cite{konig2022weakly}; (l) UP-CrackNet \cite{ma2024up}; (m) WP-CrackNet.}
		\label{fig4}
	\end{center}
\end{figure*}

\subsection{Comparison with SoTA Methods}

To validate the effectiveness of our proposed WP-CrackNet, we conduct a comprehensive comparison against four SoTA general weakly-supervised semantic methods, five supervised, one weakly-supervised and one unsupervised methods specifically designed for road crack detection. The evaluation is performed on the Crack500 \cite{yang2019feature}, Deepcrack \cite{liu2019deepcrack}, and CFD \cite{shi2016automatic} datasets. Quantitative and qualitative results are presented in Table \ref{table_c1}, Table \ref{table_c2}, Table \ref{table_c3} and Fig. \ref{fig4}, respectively. The results clearly indicate that WP-CrackNet achieves detection performance comparable to road crack detection-specific supervised methods while utilizing only image-level labels, outperforming other weakly-supervised methods as well as the unsupervised method.

Specifically, across the three datasets, our proposed WP-CrackNet demonstrates an improvement in IoU of $12.012\%-41.633\%$, $7.644\%-38.569\%$, and $31.880\%-35.716\%$ compared to other SoTA general and road crack detection-specific weakly-supervised methods. 
These results stem from WP-CrackNet’s online and iterative pseudo-label generation with an selection mechanism, which reduces noise and enhances label reliability, as well as the incorporation of CECCM and PAAM. CECCM enables more precise crack CAMs, while PAAM effectively fuses high-level semantics from the classifier with low-level structural cues from the reconstructor by modeling spatial and channel-wise dependencies. Furthermore, compared with the road crack detection-specific method WS-SCS, WP-CrackNet reduces reliance on hand-crafted cues and adopts an end-to-end joint optimization strategy, leading to greater adaptability and robustness.

When compared to five supervised methods tailored for road crack detection, WP-CrackNet exhibits only a marginal IoU drop of $0.313\%$ and $2.420\%$ compared to CT-crackseg \cite{tao2023convolutional} and outperforms other supervised methods on the DeepCrack and CFD datasets. On the Crack500 dataset, it experiences an IoU decrease of $4.494\%-9.085\%$ relative to these supervised methods. The relatively lower performance on the Crack500 dataset can be attributed primarily to two factors: (1) Higher scene variability and diverse crack morphologies: Crack500 contains four crack types with variations in widths, lengths, and branching patterns, captured on different road materials under diverse conditions. Real-world challenges such as shadows, occlusions, and lighting variations, together with complex background textures, increase intra-class variability and make crack boundary localization more difficult when only image-level labels are available; (2) More pixel-level labels for supervised methods: Crack500 contains a larger number of finely annotated pixel-level labels than the other datasets, allowing supervised networks to learn precise geometric priors and handle small or partially occluded cracks more effectively. In contrast, WP-CrackNet relies on implicit localization through image-level cues, which limits its boundary accuracy in these cases. Nevertheless, considering the substantial reliance of supervised techniques on extensive pixel-level manual annotations, our proposed WP-CrackNet, which achieves comparable detection results using only image-level labels, offers significant potential to enhance the scalability of road defect detection.

Furthermore, in comparison to the unsupervised method UP-CrackNet \cite{ma2024up}, WP-CrackNet achieves improvements in IoU by $0.899\%$, $10.410\%$, and $27.851\%$ on the Crack500, DeepCrack, and CFD datasets, respectively. The results align with the intuitive expectation that leveraging image-level label information generally leads to better detection performance than methods that do not utilize any label information. Notably, for datasets such as CFD, where cracks are thin and not obvious, our proposed WP-CrackNet exhibits superior accuracy in crack boundary prediction while effectively suppressing noise in the detection results.

\begin{table}[t!]
	\renewcommand{\arraystretch}{1}
	\settablefont
	\caption{Quantitative experimental results
    in terms of model parameters and processing efficiency.}
	\centering
        \footnotesize
	\begin{tabular}{C{2.5cm}|C{2.5cm}C{2.3cm}}
		\toprule
		\raisebox{-1.2ex}{Methods}  & Model Parameters~($M$)$\downarrow$ & FPS (on RTX3090)$\uparrow$    \\ \midrule
        {Deepcrack18 \cite{zou2018deepcrack}} & 30.905 & 59.175  \\
        {Deepcrack19 \cite{liu2019deepcrack}} & 14.720  & 287.888  \\
        {SCCDNet \cite{li2021sccdnet}} & 31.705  & 143.585  \\
        {Crack-Att \cite{xu2023crack}} & 45.804  & 64.710  \\
        {HrSegNet\cite{li2023real}} & 9.641  & 285.635 \\
        {CDLN \cite{manjunatha2024crackdenselinknet}} & 19.151  & 67.984 \\ 
        {LECSFormer \cite{chen2022refined}} & 16.528 & 96.881 \\
        {CT-crackseg \cite{tao2023convolutional}} & 22.882 & 36.890 \\
        \midrule
        {WP-CrackNet} & 86.458 & 73.613 \\
		\bottomrule
	\end{tabular}
	\label{tabel_FPS}
\end{table}

Table \ref{tabel_FPS} reports model parameters and processing efficiency of WP-CrackNet and representative supervised road crack detection methods on an NVIDIA RTX3090 using $256\times256$ inputs. For WP-CrackNet, in the inference phase, only the trained $Rec.E$, $Cls$, $Det$, and the PAAM are required for processing the test data. Experimental results show that WP-CrackNet has a slightly larger number of parameters compared with these supervised methods, and its FPS ranks at a moderate level. The relatively larger parameter size of WP-CrackNet is mainly attributed to the multi-module collaborative training strategy designed to achieve weakly supervised road crack detection. Considering that WP-CrackNet requires only image-level annotations while delivering detection performance comparable to supervised methods, it remains highly practical. In future iterations, techniques such as model pruning and knowledge distillation will be employed to compress the parameter size, along with hardware-friendly designs, to make the model suitable for deployment on edge devices (such as drones) for real-time road crack detection tasks.

\section{Conclusion}
\label{sec.conclusion}

In this paper, we propose WP-CrackNet, an innovative end-to-end weakly-supervised road crack detection method that leverages image-level labels to reduce reliance on costly pixel-level annotations, 
greatly enhancing the scalability of road inspection. WP-CrackNet consists of three components: a classifier that creates CAMs, a reconstructor that assesses the inferability between road and crack features, and a detector that outputs pixel-wise road crack detection results. The classifier and reconstructor are trained adversarially in turns, while the detector is trained with pseudo labels derived from post-processed crack CAMs. Our designed PAAM effectively fuses high-level semantics from the classifier and low-level structural cues from the reconstructor by modeling spatial and channel-wise dependencies, improving the detection performance. Additionally, the proposed CECCM improves the quality of crack CAMs through center Gaussian weighting and consistency constraints, optimizing pseudo-label generation. Extensive experiments conducted on three datasets demonstrate the effectiveness of WP-CrackNet and its superiority over SoTA general and road crack detection-specific weakly-supervised methods in detecting road cracks by only using image-level labels. 

Future work will focus on compressing and accelerating WP-CrackNet through model pruning and knowledge distillation, enabling not only cloud-based detection but also real-time deployment on edge devices. Additionally, we plan to incorporate a small amount of fine-grained pixel-level annotations for joint training, aiming to further improve detection performance and enhance domain adaptation capabilities.

\section{Acknowledgements}
This research was supported by the National Natural Science Foundation of China under Grant 62473288, the Fundamental Research Funds for the Central Universities, NIO University Programme (NIO UP), and the Xiaomi Young Talents Program.

\bibliographystyle{elsarticle-num} 
\bibliography{main}

\vspace{2em} 

\begin{wrapfigure}{l}{0.28\columnwidth} 
    \vspace{-10pt} 
    \includegraphics[width=\linewidth]{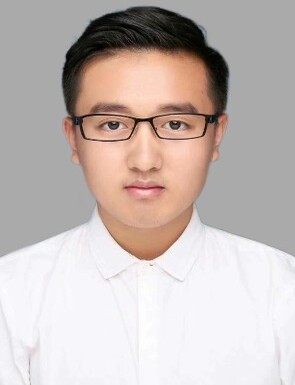}
\end{wrapfigure}
\noindent
\small
\textbf{Nachuan Ma} received the B.E.\ degree in Electrical Engineering and Automation 
    from China University of Mining and Technology in 2019, and the M.Sc.\ degree in 
    Electronic Engineering from The Chinese University of Hong Kong in 2020. He worked 
    as a Research Assistant at the Department of Electronic and Electrical Engineering, 
    Southern University of Science and Technology, from 2020 to 2021. He is pursuing his 
    Ph.D.\ degree at Tongji University with a research focus on computer vision and deep learning.

\vspace{2.5em} 

\noindent
\begin{minipage}[t]{0.28\columnwidth}
    \vspace{0pt} 
    \includegraphics[width=\linewidth]{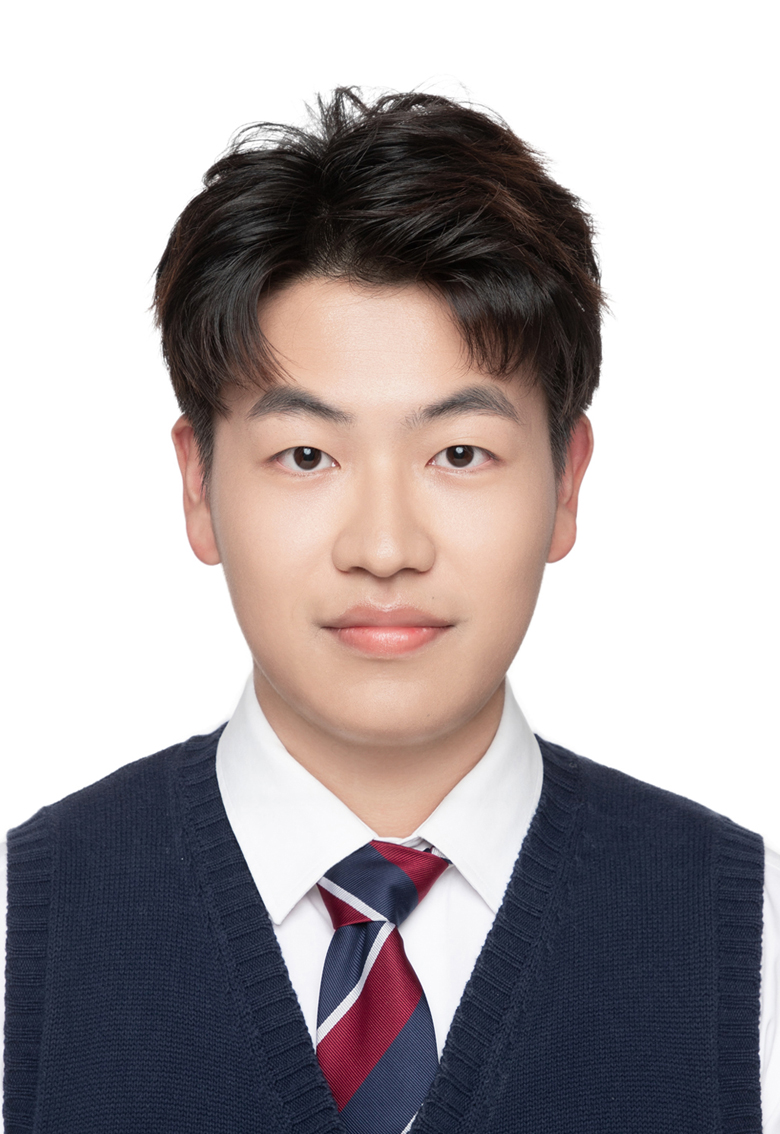}
\end{minipage}%
\hfill
\begin{minipage}[t]{0.70\columnwidth}
    \vspace{0pt} 
    \small
    \textbf{Zhengfei Song} is currently pursuing his B.E. degree at Tongji University 
    with a research focus on computer vision techniques for autonomous driving.
\end{minipage}

\vspace{3em} 

\noindent
\begin{minipage}[t]{0.28\columnwidth}
    \vspace{0pt} 
    \includegraphics[width=\linewidth]{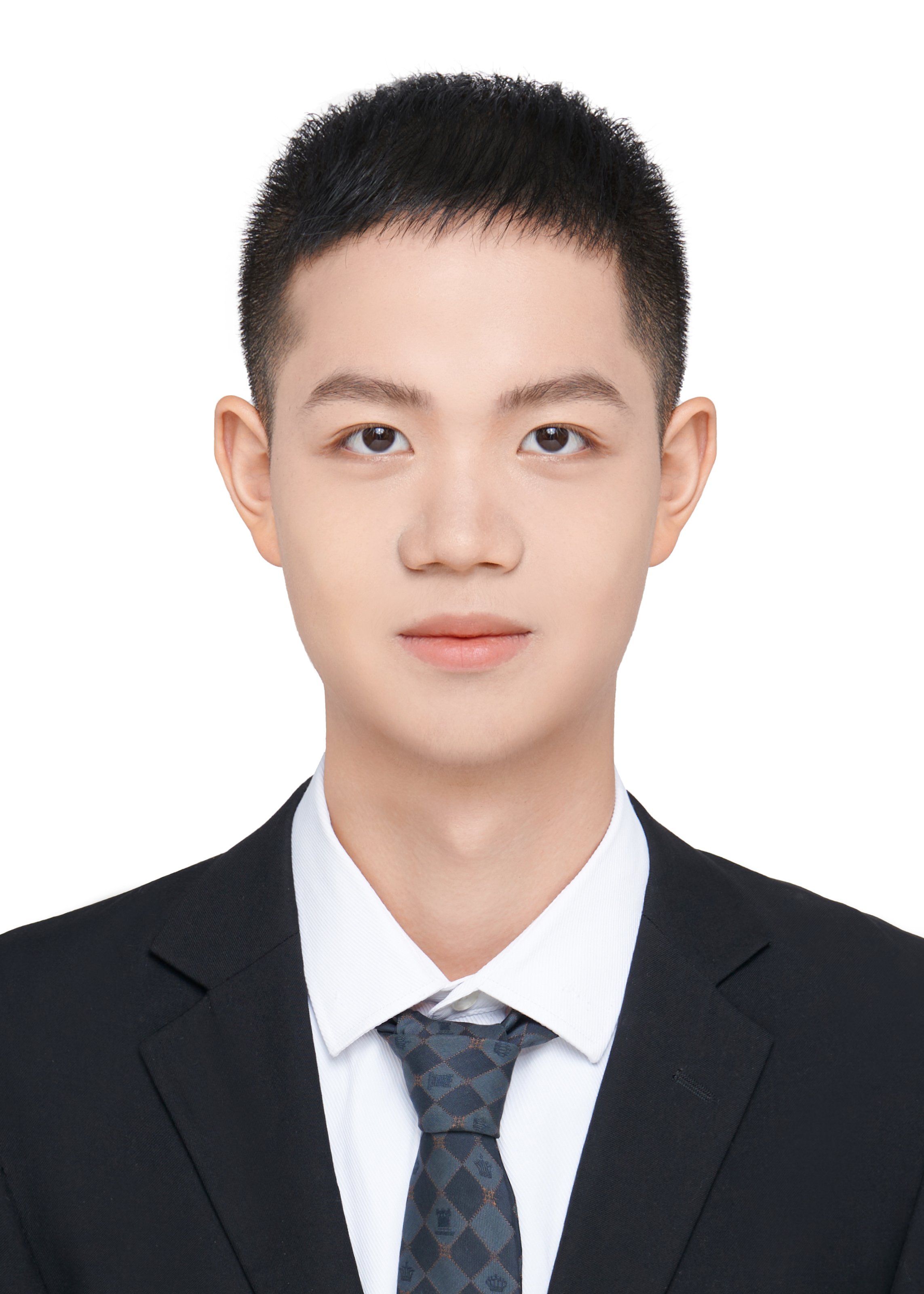}
\end{minipage}%
\hfill
\begin{minipage}[t]{0.70\columnwidth}
    \vspace{0pt} 
    \small
    \textbf{Qiang Hu} is currently pursuing his B.E. degree at Tongji University with a research focus on computer vision techniques for autonomous driving.
\end{minipage}

\vspace{4em} 

\begin{wrapfigure}{l}{0.28\columnwidth} 
    \vspace{-4pt} 
    \includegraphics[width=\linewidth]{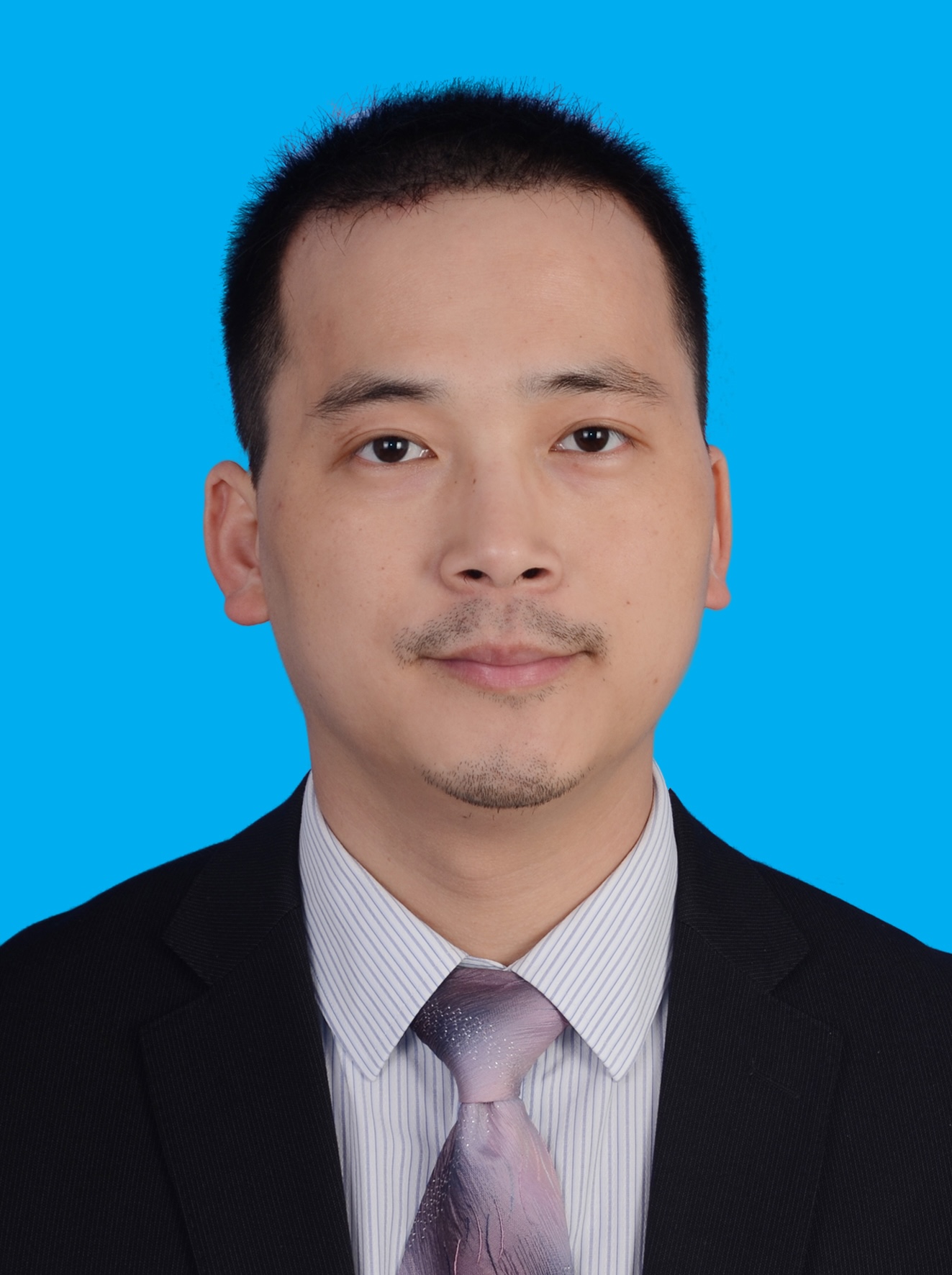}
\end{wrapfigure}
\noindent
\small
\textbf{Xiaoyu Tang} received a B.S. degree from South China Normal University, Guangzhou, China, in 2003 and an M.S. degree from Sun Yat-sen University, Guangzhou, China, in 2011. He is currently pursuing a Ph.D. degree with South China Normal University. Mr. Tang works as an associate professor, master supervisor, and deputy dean at Xingzhi College, South China Normal University. His research interests include image processing and intelligent control, artificial intelligence, the Internet of Things, and educational informatization.

\vspace{4em} 

\begin{wrapfigure}{l}{0.28\columnwidth} 
    \vspace{-8pt} 
    \includegraphics[width=\linewidth]{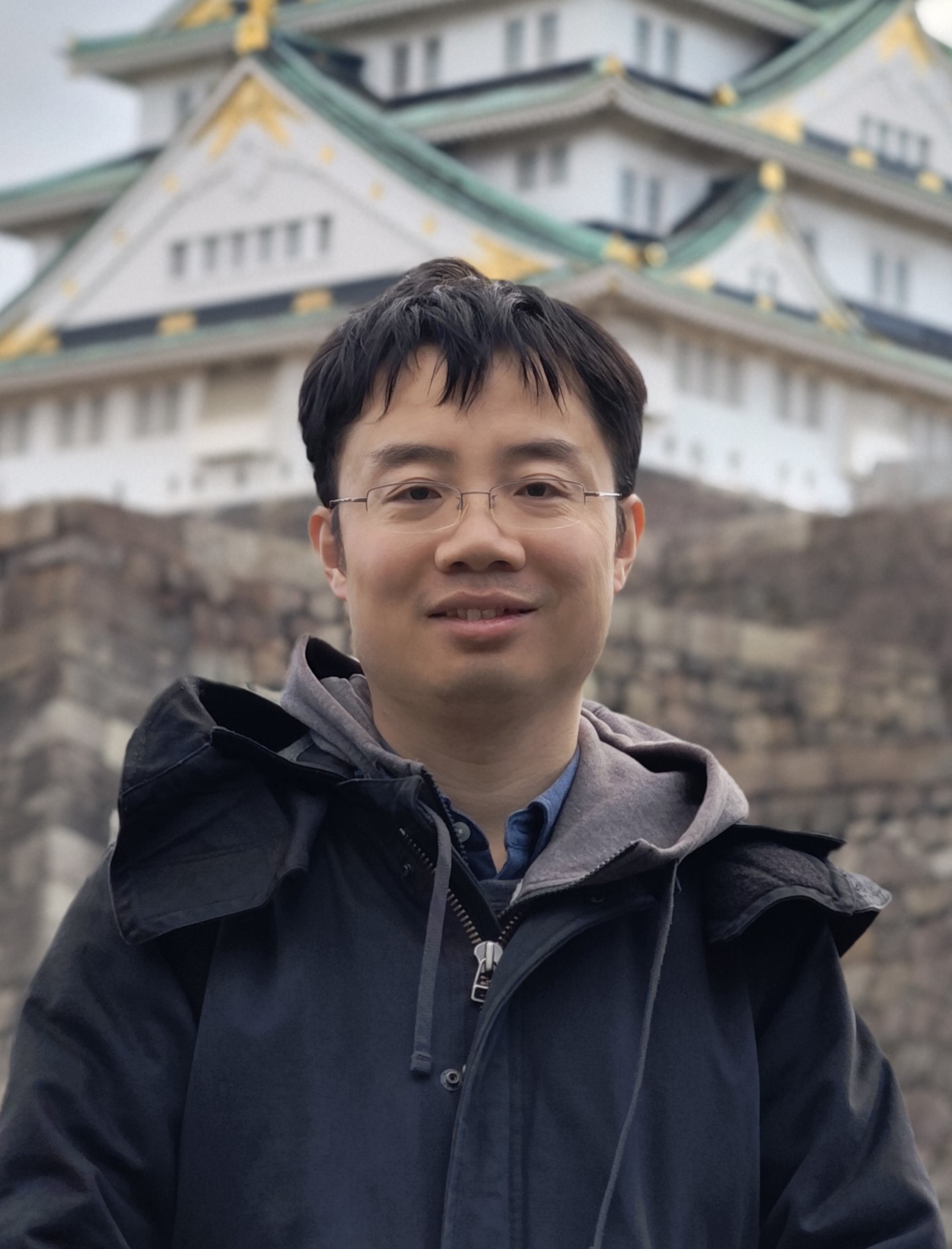}
\end{wrapfigure}
\noindent
\small
\textbf{Chengxi Zhang} received B.S. and M.S. degrees from Harbin Institute of Technology, China, in 2012 and 2015; and Ph.D. degree from Shanghai Jiao Tong University, China, in 2019. He is now an Associate Professor at Jiangnan University, China.
His interests are space engineering, robotic systems \& control. He is a Committee Member of CAA-YAC,
a Distinguished Reviewer for Intelligence \& Robotics, 2024. He is an Associate Editor of Frontiers in Aerospace Engineering, on the Editorial Board of Aerospace Systems and Astrodynamics.

\vspace{4em} 

\begin{wrapfigure}{l}{0.28\columnwidth} 
    \vspace{-8pt} 
    \includegraphics[width=\linewidth]{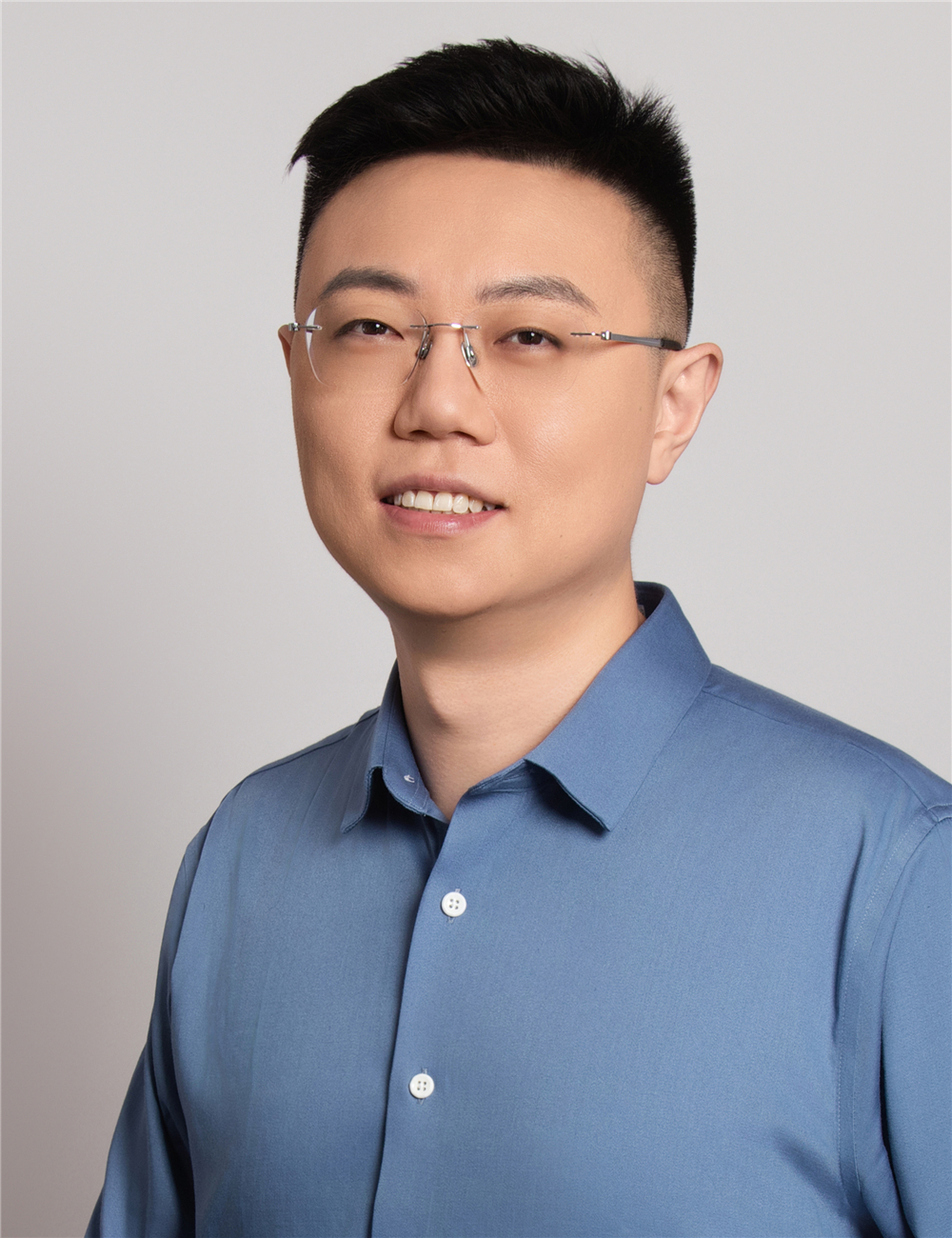}
\end{wrapfigure}
\noindent
\small
\textbf{Rui Fan} received the B.Eng. degree in Automation from the Harbin Institute of Technology in 2015 and the Ph.D. degree in Electrical and Electronic Engineering from the University of Bristol in 2018. He worked as a Research Associate at the Hong Kong University of Science and Technology from 2018 to 2020 and a Postdoctoral Scholar-Employee at the University of California San Diego between 2020 and 2021. He began his faculty career as a Full Research Professor in the College of Electronics \& Information Engineering at Tongji University in 2021. He was promoted to Full Professor in 2022 and attained tenure in 2024, both in the same college and at the Shanghai Research Institute for Intelligent Autonomous Systems. His research interests include computer vision, deep learning, and robotics, with a specific focus on humanoid visual perception under the two-streams hypothesis. Prof. Fan served as an associate editor for ICRA'23/25 and IROS'23/24, an area chair for ICIP'24, and a senior program committee member for AAAI'23/24/25/26. He organized several impactful workshops and special sessions in conjunction with WACV'21, ICIP'21/22/23, ICCV'21, ECCV'22, and ICCV'25. He was honored by being included in the Stanford University List of Top 2\% Scientists Worldwide between 2022 and 2024, recognized on the Forbes China List of 100 Outstanding Overseas Returnees in 2023, acknowledged as one of Xiaomi Young Talents in 2023, and awarded the Shanghai Science \& Technology 35 Under 35 honor in 2024 as its youngest recipient.

\vspace{4em} 

\begin{wrapfigure}{l}{0.28\columnwidth} 
    \vspace{-6pt} 
    \includegraphics[width=\linewidth]{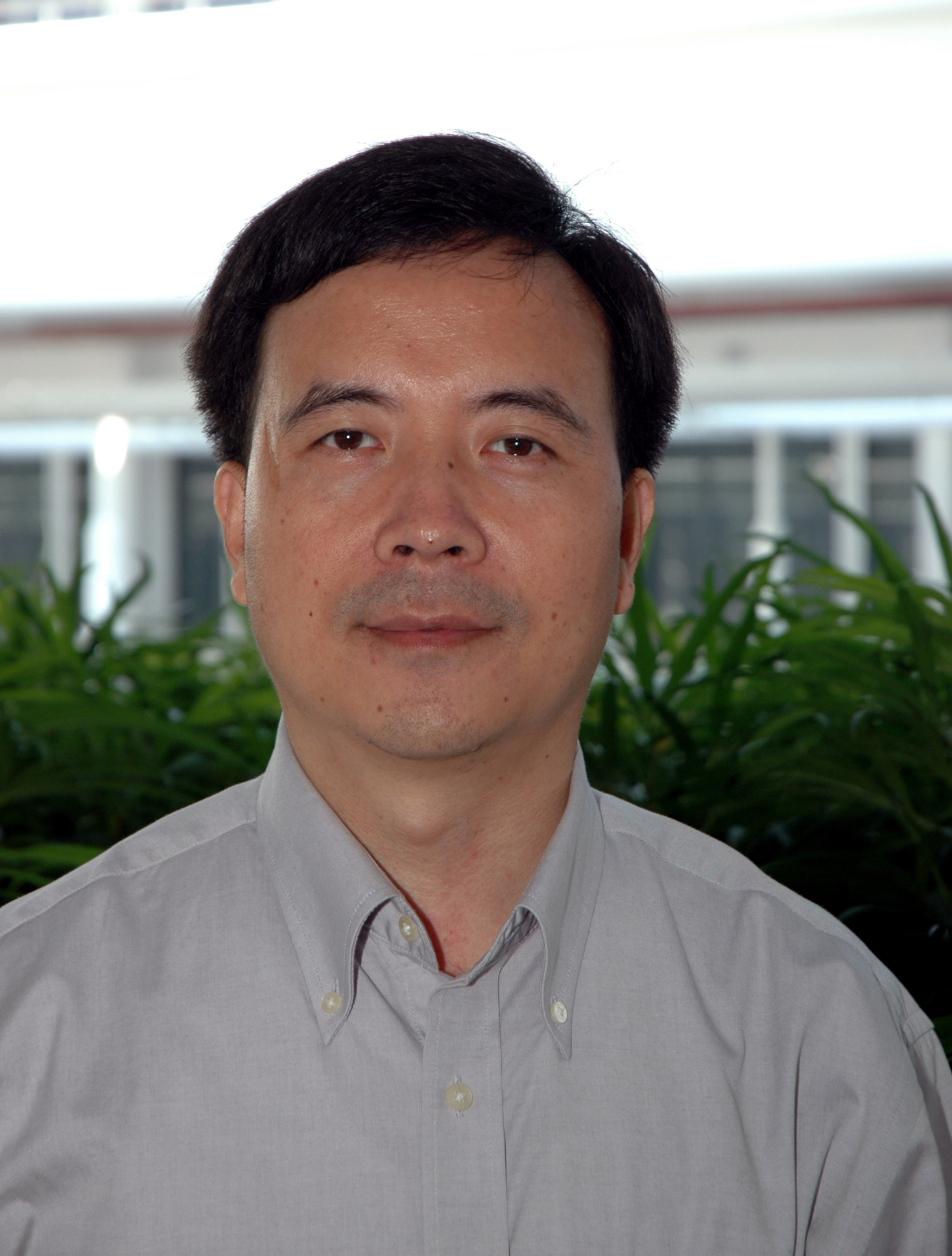}
\end{wrapfigure}
\noindent
\small
\textbf{Lihua Xie} received the Ph.D. degree in electrical engineering from the University of Newcastle, Australia, in 1992. Since 1992, he has been with the School of Electrical and Electronic Engineering, Nanyang Technological University, Singapore, where he is currently a professor and Director, Center for Advanced Robotics Technology Innovation. He served as the Head of Division of Control and Instrumentation and Co-Director, Delta-NTU Corporate Lab for Cyber-Physical Systems. He held teaching appointments in the Department of Automatic Control, Nanjing University of Science and Technology from 1986 to 1989.

Dr Xie’s research interests include robust control and estimation, networked control systems, multi-agent networks, and unmanned systems. He is an Editor-in-Chief for Unmanned Systems and Associate Editor for IEEE Control System Magazine. He has served as Editor of IET Book Series in Control and Associate Editor of a number of journals including IEEE Transactions on Automatic Control, Automatica, IEEE Transactions on Control Systems Technology, IEEE Transactions on Network Control Systems, and IEEE Transactions on Circuits and Systems-II. He was an IEEE Distinguished Lecturer (Jan 2012 – Dec 2014). Dr Xie is Fellow of Academy of Engineering Singapore, IEEE, and IFAC.

\end{document}